\documentclass[journal]{IEEEtran}

\usepackage{amsmath}
\usepackage{amssymb}
\usepackage{booktabs}
\usepackage{multirow}
\usepackage{array}
\usepackage{float}
\usepackage{graphicx}
\usepackage{cite}
\usepackage{hyperref}

\usepackage{multirow} 
\usepackage{multicol} 
\usepackage{wrapfig} 
\ifCLASSINFOpdf
\else
\fi
\begin{document}
%
\title{Multi-dimensional Visual Prompt Enhanced Image Restoration via Mamba-Transformer Aggregation}
%
%
%

\author{Aiwen~Jiang~\IEEEmembership{Member,~IEEE,}
        Hourong Chen,
        Zhiwei Chen,
        Jihua Ye,
        Mingwen Wang~\IEEEmembership{Member,~IEEE}
\thanks{Aiwen Jiang, Zhiwen Chen, and Mingwen Wang are with the School of Digital Industry, Jiangxi Normal University, Shangrao,
330004 China e-mail: jiangaiwen@jxnu.edu.cn.}
\thanks{Hourong Chen, and Jihua Ye are with School of Computer and Information Engineering, Jiangxi Normal University.}
\thanks{Manuscript received XX. XX, 2024; revised XX XX, XXXX.}}

\markboth{Journal of \LaTeX\ Class Files,~Vol.~XX, No.~XX, XXXX~XXXX}%
{Shell \MakeLowercase{\textit{et al.}}: Bare Demo of IEEEtran.cls for IEEE Journals}

\maketitle

\begin{abstract}
Image restoration is an important research topic that has wide industrial applications in practice. Traditional deep learning-based methods were tailored to specific degradation type, which limited their generalization capability. Recent efforts have focused on developing "all-in-one" models that can handle different degradation types and levels within single model. However, most of mainstream Transformer-based ones confronted with dilemma between model capabilities and computation burdens, since self-attention mechanism quadratically increase in computational complexity with respect to image size, and has inadequacies in capturing long-range dependencies. Most of Mamba-related ones solely scanned feature map in spatial dimension for global modeling, failing to fully utilize information in channel dimension. To address aforementioned problems, this paper has proposed to fully utilize complementary advantages from Mamba and Transformer without sacrificing computation efficiency. Specifically, the selective scanning mechanism of Mamba is employed to focus on spatial modeling, enabling capture long-range spatial dependencies under linear complexity. The self-attention mechanism of Transformer is applied to focus on channel modeling, avoiding high computation burdens that are in quadratic growth with image's spatial dimensions. Moreover, to enrich informative prompts for effective image restoration, multi-dimensional prompt learning modules are proposed to learn prompt-flows from multi-scale encoder/decoder layers, benefiting for revealing underlying characteristic of various degradations from both spatial and channel perspectives, therefore, enhancing the capabilities of "all-in-one" model to solve various restoration tasks. Extensive experiment results on several image restoration benchmark tasks such as image denoising, dehazing, and deraining, have demonstrated that the proposed method can achieve new state-of-the-art performance, compared with many popular mainstream methods. Related source codes and pre-trained parameters will be public on github \url{https://github.com/12138-chr/MTAIR}.
\end{abstract}


\IEEEpeerreviewmaketitle

\section{Introduction}
In real world, adverse weather conditions (such as haze and rain), as well as imperfections in imaging systems and transmission media often lead to image quality degradation. The degradations manifest as reduced sharpness, blurred details, weakened contrast, and increased noise etc. In practice, image degradation can seriously interfere with effective execution of intelligent vision system. Therefore, the restoration of high-definition and visually pleasing clear images from damaged or low-quality images has become important research topic with excellent academic and industrial application value.

Traditionally, specific deep learning-based image restoration methods were trained for specific task, such as image denoising\cite{torun2024hyperspectral,tian2023multi,wang2022blind2unblind,zhao2022hybrid}, image dehazing\cite{yi2021efficient,xiao2024single,kumari2024new,del2022new}, and image deraining\cite{ozdenizci2023restoring,chen2023learning,peng2021ensemble,peng2020cumulative,wang2022self,yi2021structure,yan2023cascaded}. These methods performed individually well in handling specific image degradation case. In practice, low-quality image often simultaneously involves multiple degradation types. Parallel deploying multiple different task-specific restoration models in single application inevitably increases computational demands and memory resources.

Recent research has begun to explore unified models that can handle multiple image degradation problems simultaneously. Methods in this category were referred as "all-in-one" models. Typically, AirNet\cite{li2022all} introduced contrastive learning and degradation-aware encoder to address the unified restoration task. Component-oriented two-stage framework IDR\cite{zhang2023ingredient} proposed to progressively restored image based on underlying physical properties that were collected for concerned degradation types. Recent works such as ProRes\cite{ma2023prores} and PromptIR\cite{potlapalli2024promptir} had introduced prompt learning into "all-in-one" task. They use learnable visual prompts to implicitly learn degradation-aware image features. They have pioneered the work revealing great potential of prompt learning in low-level image restoration field.

Most of mainstream image restoration methods were purely Transformer-based \cite{vaswani2017attention} framework. However, due to quadratic increase in computational complexity of self-attention mechanism with respect to image size, and inadequacies in capturing long-range dependencies\cite{carion2020end,dosovitskiy2020image,liu2021swin}, they have confronted with great challenge, facing dilemma between model capabilities and computation burdens.

Recently, State Space Models (SSMs)\cite{gu2023mamba,gu2021combining,smith2022simplified} has shown significant advantages in long sequence modeling in natural language processing (NLP) tasks compared to Transformers, while having linear computational complexity. Specifically, Mamba model\cite{gu2023mamba}, which has selective scanning mechanism and efficient hardware design, has been successfully surpassed Transformer on many computer vision tasks\cite{liu2024vmamba,ma2024u,zhu2024vision}. However, there is a fly in the ointment. These methods solely scanned image feature map in spatial dimension for global modeling, failing to fully utilize information in channel dimension.

We believe that multi-dimensional characteristics cannot be ignored for comprehensive image modeling. To address aforementioned problems, this paper has proposed to fully utilize complementary advantages from Mamba and Transformer without sacrificing computation efficiency. Specifically, we employed the selective scanning mechanism of Mamba to focus on spatial modeling, enabling capture long-range spatial dependencies under linear complexity. We employed the self-attention mechanism of Transformer to focus on channel modeling, avoiding high computation burdens that are in quadratic growth with image's spatial dimensions. We call the proposed method as MTAIR (Image Restoration via Mamba-Transformer Aggregation).

Moreover, to enrich informative prompts for effective image restoration, we have further designed Spatial-Channel Prompt Blocks (S-C Prompts) as prompt learning modules in multi-scale stages. Different from traditional preset prompts, herein our learned prompt flows are more multidimensional, capable of better revealing underlying characteristic of various degradations for "all-in-one" image restoration task.

In summary, the main contributions are as followings:
\begin{itemize}
    \item We have proposed a new state-of-the-art "all-in-one" image restoration method based on Mamba-Transformer cross-dimensional collaboration. In the proposed method, selective scanning mechanism in Mamba serves for long-range dependencies modeling in spatial dimension, while self-attention mechanism in Transformer serves for discriminative feature learning in channel dimension. As a result, complementary advantages from Mamba and Transformer can be fully utilized within restricted computation resource.
    \item We have designed a novel multi-dimensional prompt learning module in the proposed method. It can learn prompt-flows from multi-scale layers, benefiting for revealing underlying characteristic of various degradations from both spatial and channel perspectives, therefore, enhancing the capabilities of "all-in-one" model to solve various restoration tasks. Additionally, the prompt learning module is plug-and-play, easy to be integrated into any other existing networks.
    \item Extensive experimental results on several image restoration benchmark tasks such as image denoising, dehazing, and deraining, have demonstrated that the proposed method can achieve new state-of-the-art performance, compared with many popular mainstream methods.
\end{itemize}

\section{Related work}
\subsection{Multi-degradation Image Restoration}
Although single-degradation image restoration had made significant progress, multi-degradation image restoration (also known as "all-in-one" image restoration) was still a challenging computer vision task. Compared with single-degradation models, multi-degradation recovery is more applicable in terms of computational demands and memory resources. Therefore, in this section, we concentrated on briefly introducing representative multi-degradation restoration work.

To address the image degradation caused by adverse factors (such as rain, fog, snow, noise), researchers have proposed various excellent multi-degradation image restoration methods. In early work, Li et al.\cite{li2020all} developed an integrated restoration model, in which dedicated encoders were respectively proposed for each degradation type, along with a shared generic decoder. Chen et al.\cite{chen2021pre} proposed image processing transformer model, which consisted of multi-head and multi-tail for different tasks and a shared transformer body including encoder and decoder.

In following work, many methods proposed to remove the complex multi-head and multi-tail structures, opting for a single-branch end-to-end network. Typically, Li et al.\cite{li2022all} utilized contrastive learning to extract various degradation representations to help single-branch network address multiple degradations. Chen et al.\cite{chen2022learning} trained a unified model based on knowledge distillation for multiple restoration models. Zhang et al~\cite{zhang2023ingredient} proposed a two-stage framework IDR, collecting task-specific knowledge on underlying physical properties of different degradation types to help gradually restore images.

In more recent, prompt-learning-based methods have been introduced into image restoration field. ProRes\cite{ma2023prores} proposed to integrate learnable visual prompt into the restoration network, while PromptIR\cite{potlapalli2024promptir} proposed learnable prompts between each level of decoders in restoration network. They utilized visual prompts to encode degradation-specific information to dynamically adjust feature representations for various degradation restoration tasks.

However, these aforementioned methods were all purely based on Transformer-based deep learning framework. Although Transformer models are superior to convolutional neural networks in capturing global dependencies and modeling complex relationships, due to the quadratical complexity of Transformer's self-attention mechanism with input size, the model scalability was largely constrained, especially in resource-limited environments or when dealing with high-resolution images in "all-in-one" tasks.

\subsection{State-Space Models}
State-space models (SSMs), inspired by classical control theory, have recently demonstrated strong competitiveness in state-space transformation domain, offering new perspectives for addressing long-range dependency problems~\cite{smith2022simplified,gu2021combining}.

Structured State-data Sequence model (S4)~\cite{gu2021efficiently} was a pioneering deep state-space work. It introduced diagonal-structured parameter normalization, providing an effective alternative to CNNs and Transformers for modeling long-range dependencies. Subsequent advancements have appeared in the form of S5~\cite{smith2022simplified}, which was built on S4 and introduced efficient parallel scanning strategies to further enhance model's performance. Gated state-space layers~\cite{mehta2022long} integrated additional gating units to enhance model's expressiveness.

In most recent, a data-dependent SSM layer and a universal language model backbone called Mamba\cite{gu2023mamba} has been proposed. It has not only outperformed Transformers on large-scale real-world datasets but also shown effectiveness and scalability with linear complexity to input sequence length. Several variants of Mamba have also been successfully applied to vision tasks such as image classification~\cite{liu2024vmamba,zhu2024vision}, video generation~\cite{wang2023selective,islam2023efficient,nguyen2022s4nd}, and biomedical image segmentation~\cite{ma2024u}, which demonstrates its broad applicability and potential in different domains.

However, when processing image data, ordinary state-space models only model data in single direction, leading to deficiencies in multi-direction perception. Although Vmamba\cite{liu2024vmamba} and Vision Mamba\cite{zhu2024vision} had proposed to perform bidirectional scanning for image data in both forward and backward directions, they still ignored information in channel dimension. Therefore, it is of significance and meaningful to design a framework that can comprehensively capture and effectively leverage multi-dimensional information from data stream, which is as well the goal of this work.

\subsection{Prompt Learning}
Prompt learning is an useful paradigm that was originated in natural language processing(NLP) field~\cite{sarkar2019text,sood2020improving}. It has achieved significant success in leveraging large language models (LLMs). In NLP, prompt learning can enhance model's performance through providing contextual information to fine-tune LLMs to be more adaptable to specific tasks.

In recent, prompt learning has been widely applied in various vision tasks~\cite{sarkar2019text,sood2020improving,wang2023imagen,xie2023smartbrush}. In the field of multi-degradation image restoration, models based on learnable prompts~\cite{ma2023prores,potlapalli2024promptir} proposed to encode degradation-specific information through learning data distribution. These learnable prompts were employed to dynamically guide restoration network for specific degradations, allowing model with efficient adaptation~\cite{zhou2022learning}.

However, the aforementioned available visual prompt-based methods constrained themselves on single dimension and fixed scale. Herein, in this paper, we proposed to learn degradation knowledge through multi-scale and multi-dimensional visual prompts for different low-level vision tasks.

\section{Method}
\begin{figure*}[!t] 
\centering 
\includegraphics[width=1.0\textwidth]{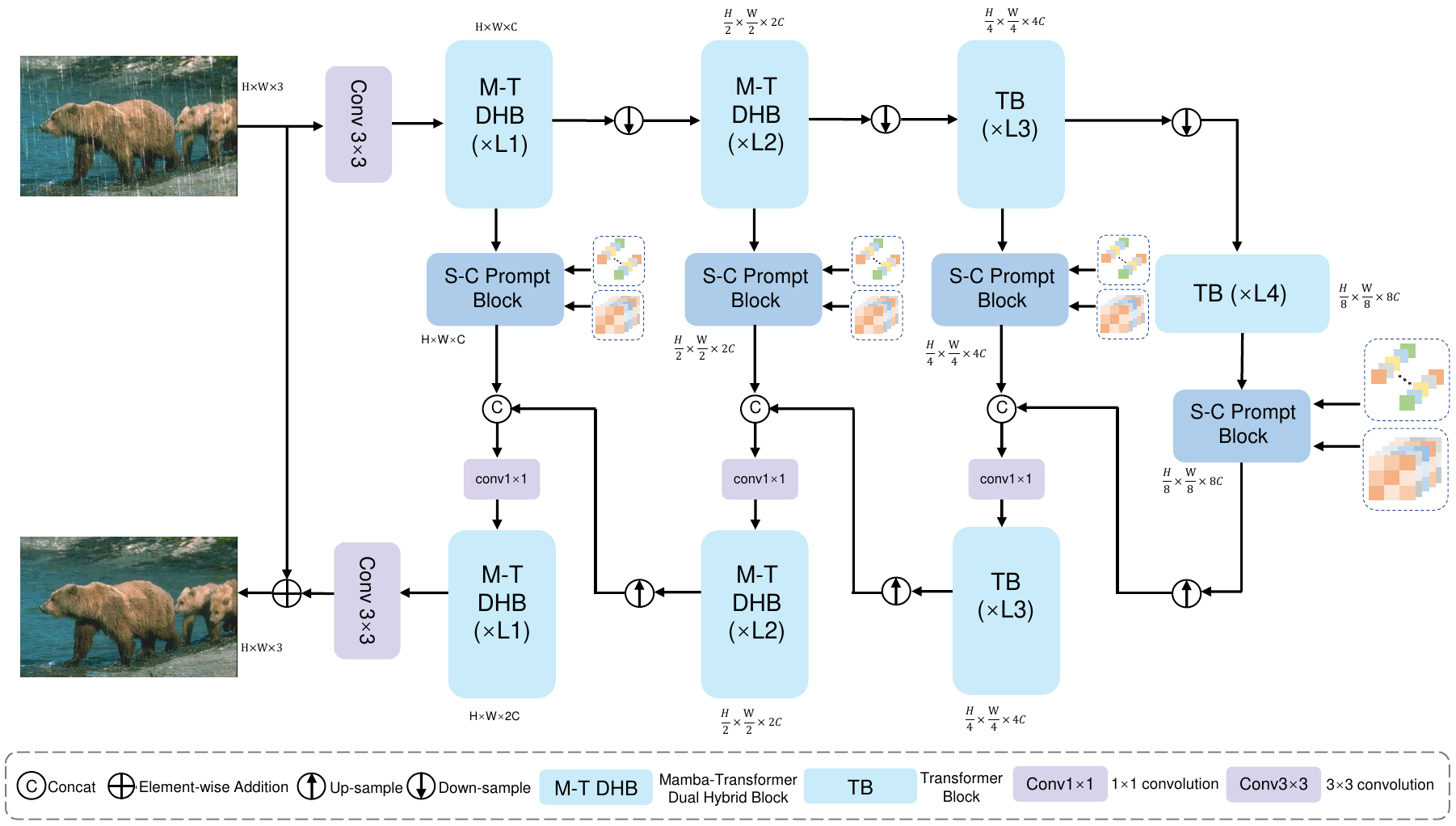} 
\caption{Overview of the MTAIR. It consists of a multi-stage encoder-decoder network(M-T DHB or TB.) and multi-stage S-C Prompt Block.} 
\label{fig:network} 
\end{figure*}

\subsection{Overall Pipeline}
In this section, we provide a preliminary introduction to our proposed MTAIR network.

The overall pipeline is illustrated in Figure~\ref{fig:network}. It is a multi-scale encoder-decoder network with skip connections by Spatial-Channel Prompt Blocks (S-C PB) across encoder-decoder layers at different level. The S-C Prompt Block is our specifically designed module for prompt-flow learning, dynamically aggregating informative degradation properties at different scales for restoration. Concatenation operation followed by a $1\times1$ bottleneck convolution is employed in skip connection, helping maintain structural and textural image details and retain informative channel features during restoration.

To avoid excessive growth of model parameters and effectively preserve informative visual details in spatial domain, MTAIR network is constructed by four encoder layers and four decoder layers. Mamba-Transformer Dual Hybrid Block (M-T DHB) is specifically designed to extract texture features in the first two shallow layers in MTAIR network. In deeper layers, basic Transformer Blocks (TB) from Restormer~\cite{zamir2022restormer} are stacked in a manner similar to U-Net configuration.

\begin{figure*}[!t] 
\centering 
\includegraphics[width=1.0\textwidth]{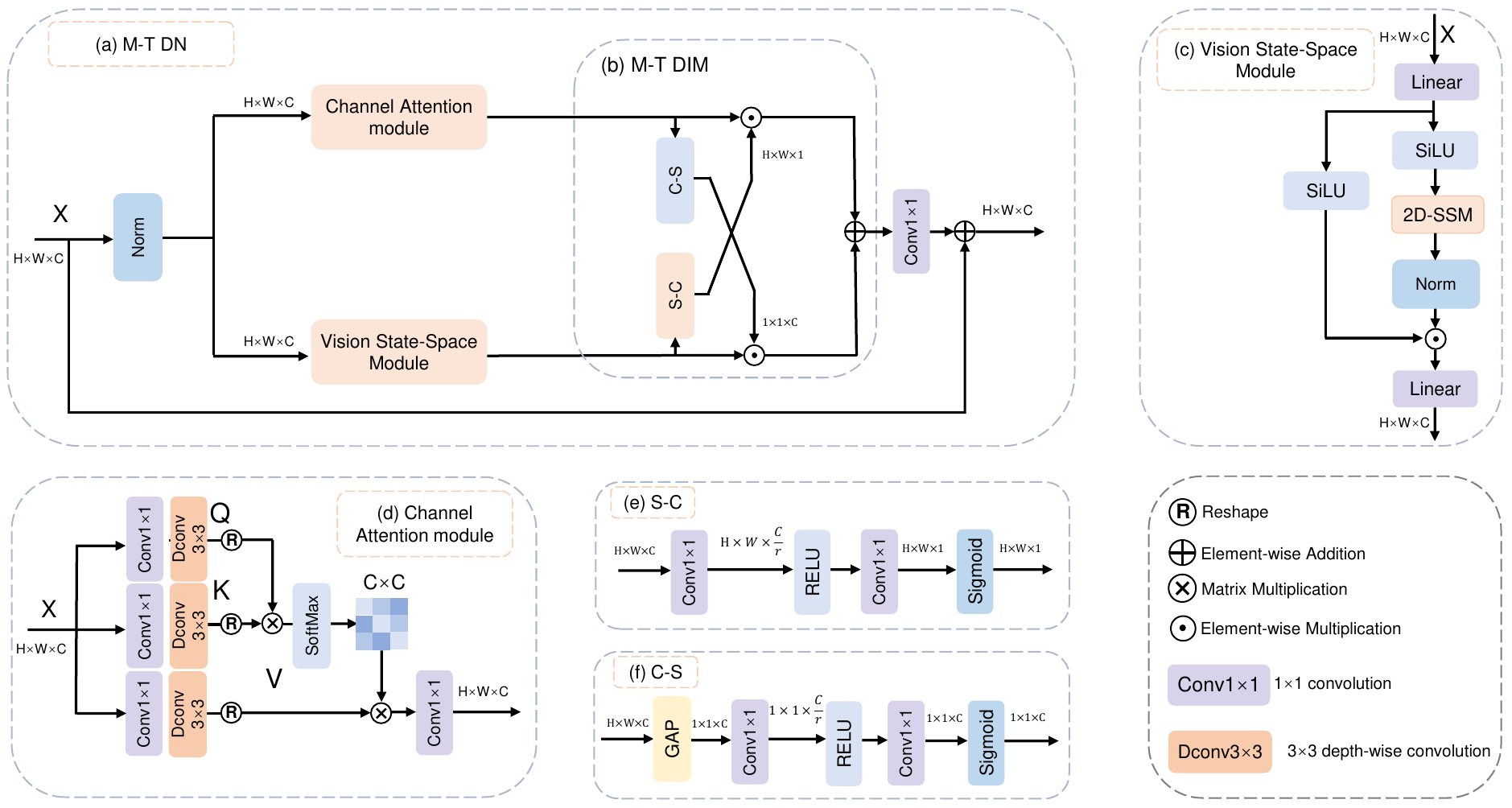} 
\caption{Overview of the M-T DHB.(a) M-T DN:Mamba-Transformer Doublebranch Network. (b) M-T DIM:Mamba-Transformer Dual Interaction Module. (c) Vision State-Space Module. (d) Channel Attention module. (e) S-C module. (f) C-S module. } 
\label{fig:M-T} 
\end{figure*}

As illustrated in Figure~\ref{fig:M-T}, the M-T DHB consists of three key components which are M-T DN (Mamba-Transformer Double-branch Network), M-T DIM (Mamba-Transformer Dual-Interaction Module), and GDFN (Gated Dconv Feed-Forward Network)~\cite{zamir2022restormer}. The details on their structures will be further introduced in subsequent sections.

Concretely, given a degraded image \( I \in \mathbb{R}^{H \times W \times 3} \), a $3 \times 3$ convolution layer is firstly applied to extract low-level feature maps \( F_0 \in \mathbb{R}^{H \times W \times C} \) from image \( I \), where \( H \times W \) represents spatial dimensions and \( C \) is channel size. Each encoder/decoder layer employs multiple M-T DHB or TB, with the \( i^{th}\) encoder/decoder layer consisting of \( L_i \) M-T DHB or TB. The number of M-T DHB or TB stacked in each layer increases progressively to ensure computation efficiency. At the same time, the channel sizes of feature maps are gradually increased while corresponding spatial resolutions are gradually reduced in encoder layers at different scale. Pixel-shuffle operations are performed for feature downsampling between layers. Ultimately latent representation \( F_l \in \mathbb{R}^{(H/8) \times (W/8) \times 8C} \) is produced in the last encoder layer. Afterward, at decoder stage, in contrary, decoder layers gradually restore the latent features \( F_l \) back to high-resolution features \( F_d \in \mathbb{R}^{H \times W \times C} \). Pixel-unshuffle operations are performed for feature upsampling. Finally, a $3 \times 3$ convolution layer is employed to map $F_d$ back to image \( \hat{I} \in \mathbb{R}^{H \times W \times 3} \) as clear output.


In the following subsections, we will describe aforementioned proposed modules in details.

\begin{figure}[h]
    \centering
    \includegraphics[width=0.5\textwidth]{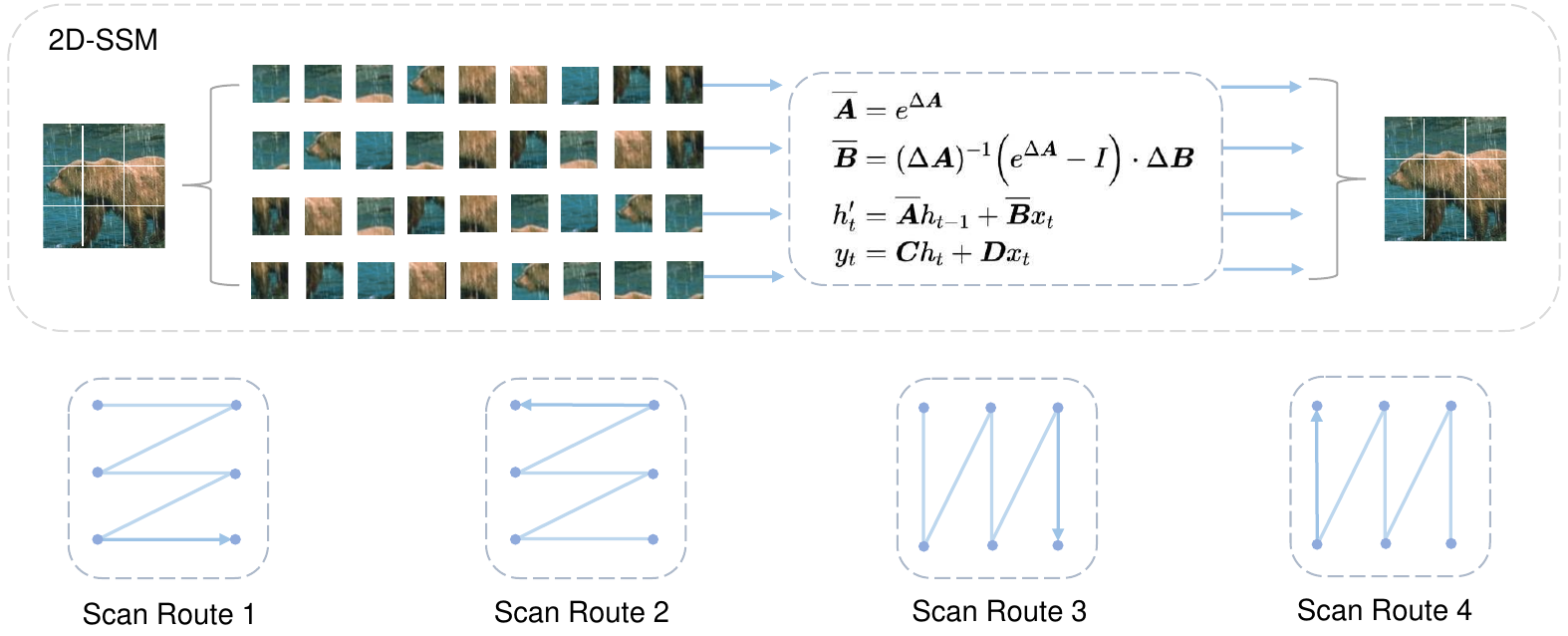} 
    \caption{The scanning route consist of four directions: from the top-left to the bottom-right, from the bottom-right to the top-left, from the top-right to the top-left, and from the bottom-left to the top-right.}
    \label{fig:Scan route}
\end{figure}

\subsection{Mamba-Transformer Dual-branch Network}
As shown in Figure \ref{fig:M-T}, the Mamba-Transformer Dual-branch Network consists of two branches. One branch applies multi-head self-attention with separable convolutions to extract channel features, while the other branch employs Vision State-Space Module with linear complexity to extract spatial features.

Specifically, given an image feature map \( X \in \mathbb{R}^{H \times W \times C} \), layer normalization\cite{ba2016layer} is first performed to obtain \( X_0 \in \mathbb{R}^{H \times W \times C} \).

In the self-attention branch, \( X_0 \) is projected to queries \( Q = W_{d}^{Q} W_{s}^{Q} X_0 \), keys \( K = W_{d}^{K} W_{s}^{K} X_0 \), and values \( V = W_{d}^{V} W_{s}^{V} X_0 \), where \( W_{s}^{(\cdot)} \) are $1 \times 1$ pointwise convolutions, and \( W_{d}^{(\cdot)} \) are $3 \times 3$ depthwise convolutions. Then, dot-product operation is performed between reshaped \( \hat{Q} \in \mathbb{R}^{HW\times \hat{C}} \) and \( \hat{K} \in \mathbb{R}^{\hat{C} \times HW}\), generating a transposed attention map \( A \in \mathbb{R}^{\hat{C} \times \hat{C}} \). Herein, $\hat{C}$ is dimension size of the projected channel feature. Compared to traditional attention map of size \( \mathbb{R}^{HW \times HW }\)\cite{dosovitskiy2020image,vaswani2017attention}, herein the transposed attention is much more efficient, since $HW \gg \hat{C}$.

Similar to traditional multi-head self-attention mechanism, herein multiple "heads" in channel direction learn separate attention maps in parallel, significantly reducing computational burden. The multi-head attention process in channel direction can be described as following Equations~\ref{eq:attention}:
\begin{equation}
\begin{aligned}
X_C = W_s (\hat{V} \cdot \text{Softmax} (\hat{K} \cdot \hat{Q} / \beta))
\end{aligned}
\label{eq:attention}
\end{equation}
where \( \beta \) is a learnable scaling parameter that controls the magnitude of dot-product between \( \hat{K} \) and \( \hat{Q} \).


In the other branch, to maintain computational efficiency while capturing long-range spatial dependencies, we introduce Mamba~\cite{gu2023mamba} scanning mechanism into image restoration. To better utilize 2D spatial information, inspired by Vmamba~\cite{liu2024vmamba}, we accept specifically the Two-Dimensional Selective Scanning strategy (2D-SSM). As shown in Figure~\ref{fig:M-T}(c), input feature \( X_0 \in \mathbb{R}^{H \times W \times C} \) is expanded to \( \hat{X_0} \in \mathbb{R}^{H \times W \times 2C} \) through a linear layer. Then, $\hat{X_0}$ is split into two portions. One portion continuously passes through a SiLU\cite{shazeer2020glu} activation, a 2D-SSM layer, and a normalization layer, resulting in feature \( X_1 \in \mathbb{R}^{H \times W \times C} \). The other portion passes through a SiLU activation, directly resulting in feature \( X_2 \in \mathbb{R}^{H \times W \times C} \). Finally, \( X_1 \) and \( X_2 \) are element-wise multiplied together, and subsequently processed through a linear layer to obtain output \( X_S\). The process is illustrated in Equations~\ref{eq:mamba ssm}.
\begin{equation}
\centering
\begin{array}{c}
X_1 = \text{Norm}(\text{2D-SSM}(\text{SiLU}(\text{linear}(X_0)))) \\
X_2 = \text{SiLU}(\text{linear}(X_0)) \\
X_S = \text{linear}(X_1 \odot X_2)
\end{array}
\label{eq:mamba ssm}
\end{equation}
where $\odot$ denotes element-wise multiplication.


In 2D-SSM layer, we perform bidirectional scanning of image features in both vertical and horizontal directions, as shown in Figure \ref{fig:Scan route}. The four directional sequences are modeled individually according to Mamba's basic unidirectional modeling strategy, and finally merged after alignment.

\begin{figure*}[t]
  \centering
  \includegraphics[width=\linewidth]{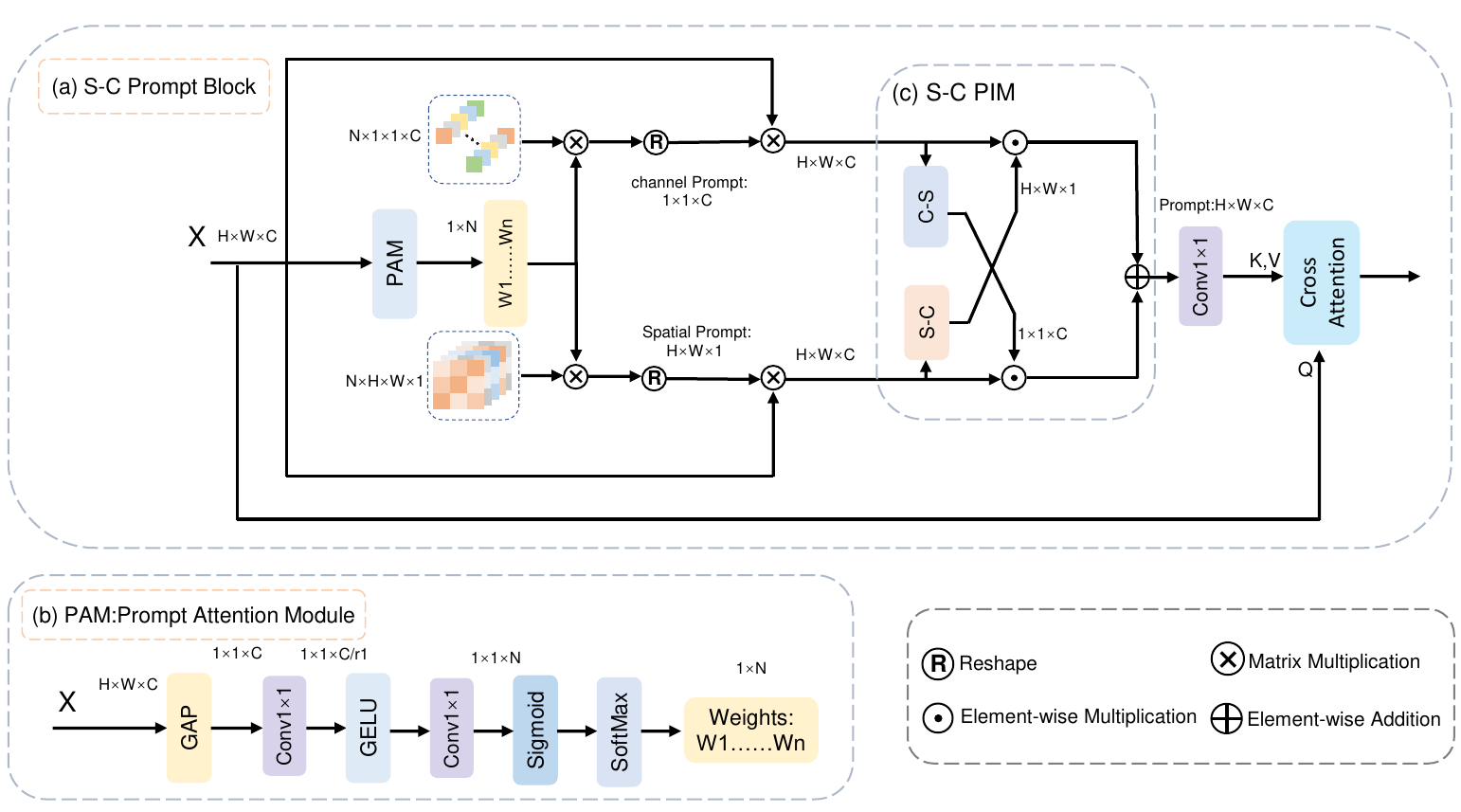}
  \caption{(a) Overview of the proposed S-C Prompt Block. (b) PAM:Prompt Attention Module. (c) S-C PIM: Spatial-Channel Prompt Interaction Module.}
  \label{fig:SC PB}
\end{figure*}

\subsection{Mamba-Transformer Dual-Interaction Module}
As shown in Figure~\ref{fig:M-T}, after M-T DN, we obtain two feature streams, one of which is the feature \(X_C\) extracted by Transformer-based attention module on channel direction, and the other one of which is the feature \(X_S\) extracted by the visual state-space module on spatial direction. Since both possess unilateral crucial information for image restoration, we propose Mamba-Transformer Dual Interaction Module (M-T DIM) deliberately for spatial-channel mutual fusion, compensating their modeling advantages for each other.

For effective cross-direction interaction, we have designed two sub-components, where "S-C" computes spatial attention map \textit{AttenS} of size $\mathbb{R}^{H \times W \times 1}$ to enrich spatial discriminative abilities for channel-branch features \(X_C \), and "C-S" computes channel attention weights \textit{AttenC} of size $\mathbb{R}^{1 \times 1 \times C}$ to enhance channel-wise discriminative for spatial-branch features \(X_S\). The computation process is as shown in following Equation~\ref{eq:MT DIM}:
\begin{equation}
\centering
\begin{array}{c}
\text{AttenC}(X_C)=\text{Sigmoid}\left(W_{2} \sigma\left(W_{1} \text{H}_\text{GAP}(X_C)\right)\right) \\
\text{AttenS}(X_S)=\text{Sigmoid}\left(W_{4} \sigma\left(W_{3} (X_S)\right)\right) \\
\hat{X_C} =X_C \odot \text{AttenS}(X_S) \\
\hat{X_S} =X_S \odot \text{AttenC}(X_C)
\end{array}
\label{eq:MT DIM}
\end{equation}
where, $\text{H}_\text{GAP}$ represents global average pooling, $\text{Sigmoid}$ represents sigmoid activation, and $\sigma(\cdot)$ represents ReLU activation. $W_i$ represents point-wise convolution weights for scaling down or up channel dimensions. The reduction ratio for $W_1$ is $r$, and the increment ratio for $W_2$ is $r$. The compression ratios of $W_3$ and $W_4$ are $r$ and $C/r$ respectively. The $\odot$ denotes element-wise multiplication. \(\hat{X_C}\) and \(\hat{X_S}\) represent the fused channel features and the fused spatial features respectively.

Finally, the two feature streams are combined together in hybrid through element-wise addition. Then, after a $1\times1$ convolution, the hybrid feature is residually connected with original feature \(X\), and output final fused feature \(\hat{X}\). The process is as shown in Equation~\ref{eq:comb-x}.
\begin{equation}
    \label{eq:comb-x}
    \hat{X} = \text{Conv}_{1 \times 1}(\hat{X_C} + \hat{X_S}) + X_0
\end{equation}

\subsection{Spatial-Channel Prompt Block}
As we know, in natural language processing field, prompt learning can adapt pre-trained large models to new tasks without extensive parameter adjustments. They use flexible, controllable, and human-understandable prompts for parameter-efficient fine-tuning. However, in low-level vision tasks, due to the complexity of degradation, it is difficult to describe degradations and their corresponding properties using language. Therefore, in the field of image restoration, we propose a learnable prompt block that effectively encodes and interacts with context information related to specific tasks. The injected prompts about degradation types and properties can guide model to adaptively adjust its latent feature distribution for corresponding restoration tasks.

Specifically, it generates a set of learnable parameters from respective spatial dimension and channel dimension, and dynamically interacts with input features to embed degradation information in both channel and spatial dimensions. Therefore, we call it Spatial-Channel (S-C) Prompt Block.

As show in Figure~\ref{fig:SC PB}, the S-C Prompt Block consists of two components. One is prompt generation module and the other one is spatial-channel prompt interaction module. The dimension sizes of the generated prompts depend on input feature. Here, we take the dimension sizes of input feature $X$ are $H \times W \times C$ for example.
\subsubsection{\textbf{Prompt Generation Module}}
The prompt generation module generates two sets of learnable parameters respectively \(P_C \in \mathbb{R}^{N \times 1 \times 1 \times C}\) on channel direction, and \(P_S \in \mathbb{R}^{N \times H \times W \times 1}\) on spatial direction, where each set owns $N$ parameter codebooks. These codebooks preserve soft visual prompts for various degradation cases.

The channel size of $P_C$ is the same as input's channel size, while the spatial size of $P_S$ is the same as input's spatial size. $N$ is closely related to the number of degradation types to be considered. In this paper, we set it 5.

To dynamically predict composite prompts from these learnable codebooks, a prompt-attention computation module (PAM) is employed to calculate the attention-based prompt weights from input feature \(X\), as shown in the figure \ref{fig:SC PB}. First, global average pooling is applied on input feature. Then, the pooled feature passes through two 1×1 convolution layers with GELU as activation function between them. Finally, Sigmoid activation and SoftMax operations are applied sequentially to produce the prompt weights \(W \in \mathbb{R}^{1 \times N}\). We utilize these prompt weights to respectively aggregate the two codebook sets, which results in composite channel prompts \(P_{C1}\) and spatial prompts \(P_{S1}\). Since the prompt weights are specifically derived from input feature \(X\), the resulted two prompts can perceive the latent discriminative information about degradations in input feature.

Overall, the aforementioned process for PAM can be summarized in Equation~\ref{eq:PAM}.
\begin{equation}
\centering
\begin{array}{c}
W =\text{Softmax}(\text{Sigmoid}\left(W_{6} \gamma\left(W_{5} H_{GAP}(X)\right)\right)) \\
P_{C1} =W \times P_C \\
P_{S1} =W \times P_S
\end{array}
\label{eq:PAM}
\end{equation}
where, $\text{H}_{\text{GAP}}$ represents global average pooling, $\text{Sigmoid}$ is sigmoid activation, $\gamma(\cdot)$ represents GeLU activation. $W_5$ and $W_6$ represent point-wise convolution weights for scaling down channel dimension.

\subsubsection{\textbf{Spatial-Channel Prompt Interaction Module}}
To dynamically adjusting feature distributions and jointly mining potential degradation properties from input feature, in Spatial-Channel Prompt Interaction Module (S-C PIM), the composite channel prompts \(P_{C1}\) and spatial prompts \(P_{S1}\) are respectively combined with input feature \(X\) through applying element-wise multiplications to obtain $\hat{P_{C1}}\in \mathbb{R}^{H \times W \times C}$ and $\hat{P_{S1}}\in \mathbb{R}^{H \times W \times C}$.

Similar to M-T DIM, the channel-prompt-guided feature $\hat{P_{C1}}$ passes through C-S module to enrich $\hat{P_{S1}}$ in channel dimension. The spatial-prompt-guided feature $\hat{P_{S1}}$ passes through S-C module to enrich $\hat{P_{C1}}$ in spatial dimension. Subsequently, the prompt-guided features $\hat{P_{C1}}$ and \(\hat{P_{S1}}\) have mutually interacted with each other, and dynamically resulted an adaptive feature-specific prompt \( \hat{P} \in \mathbb{R}^{H \times W \times C}\) after one 1×1 convolution.

Finally, multi-Dconv heads for transposed cross-attention\cite{chen2021crossvit,hertz2022prompt,rombach2022high} are employed. The feature-specific prompt \(\hat{P}\) is fused with input features \(X\) through cross-attention, emphasizing informative degradation properties both in spatial and channel dimensions. Here, the queries \(Q_f\) are derived from input features \(X\), and the keys \(K_p\) and values \(V_p\) come from feature-specific prompt \(\hat{P}\).

The aforementioned process can be summarized as following Equation~\ref{eq:S-C PIM}:
\begin{equation}
\centering
\begin{array}{c}
\hat{P_{C}} =\hat{P_{C1}} \odot \text{AttenS}(\hat{P_{S1}}) \\
\hat{P_{S}} =\hat{P_{S1}} \odot \text{AttenC}(\hat{P_{C1}}) \\
\hat{P} =Conv_{1 \times 1}(\hat{P_{C}} + \hat{P_{S}}) \\
\hat{X} =W_s (\hat{V_p} \cdot \text{Softmax} (\hat{K_p} \cdot \hat{Q_f} / \beta))
\end{array}
\label{eq:S-C PIM}
\end{equation}
where, the $\odot$ denotes element-wise multiplication.

Overall, the proposed prompt block is plug-and-play. It operates on the skip connections between encoder-decoder layers, dynamically adjusting the latent features flowing from encoder to the corresponding decoder at each level.

\section{Experiment}
To demonstrate the effectiveness of the proposed MTAIR, we have conducted extensive experiments on various datasets for three typical image restoration tasks (denoising, deraining and dehazing). In this section, we will describe the details of experimental setup, provide qualitative and quantitative analysis results, and discuss the impacts of each proposed key component in ablation studies.

\begin{table*}[!ht]
\small 
\centering
\setlength{\tabcolsep}{9.0pt} 
\renewcommand\arraystretch{1.15} 
\caption{Comparisons with state-of-the-art all-in-one image restoration methods under all-in-one restoration setting.}
\label{tab:all in one result}
\begin{tabular}{c|ccccc|c}
\hline
\multirow{2}{*}{Method} & \multicolumn{3}{|c}{ Denoising on CBSD68 dataset } & {Deraining} & {Dehazing} & \multirow{2}{*}{ Average } \\
                        &  $\sigma=15$  &  $\sigma=25$  &  $\sigma=50$  &  on Rain100L &  on SOTS & \\
\hline
  BRDNet\cite{tian2020image} & 32.26/0.898 & 29.76/0.836 & 26.34/0.693 & 27.42/0.895 & 23.23/0.895 & 27.80/0.843 \\
  LPNet\cite{fu2019lightweight} & 26.47/0.778 & 24.77/0.748 & 21.26/0.552 & 24.88/0.784 & 20.84/0.828 & 23.64/0.738 \\
  FDGAN\cite{dong2020fd} & 30.25/0.910 & 28.81/0.868 & 26.43/0.776 & 29.89/0.933 & 24.71/0.929 & 28.02/0.883 \\
  MPRNet\cite{zamir2021multi} & 33.54/0.927 & 30.89/0.880 & 27.56/0.779 & 33.57/0.954 & 25.28/0.955 & 30.17/0.899 \\
  DL\cite{fan2019general} & 33.05/0.914 & 30.41/0.861 & 26.90/0.740 & 32.62/0.931 & 26.92/0.931 & 29.98/0.875 \\
  AirNet\cite{li2022all} & 33.92/\underline{0.933} & 31.26/\underline{0.888} &28.00/0.797 & 34.90/0.968 & 27.94/0.962 & 31.20/0.910 \\
  TKMANet\cite{chen2022learning} & 33.02/0.924 & 30.31/0.820 & 23.80/0.556 & 34.94/0.972 & 30.41/0.973 & 30.50/0.849 \\
  LoRA-IR\cite{ai2024lora} & 34.06/0.935 & 31.42/0.819 & 28.18/0.803 & 37.75/0.979 & 30.68/0.961 & 32.42/0.914 \\
  DA-CLIP\cite{luo2023controlling} & 30.02/0.821 & 24.86/0.585 & 22.29/0.476 & 36.28/0.968 & 29.46/0.963 & 28.58/0.763 \\
  PromptIR\cite{potlapalli2024promptir} & \underline{33.98}/\underline{0.933} & \underline{31.31}/\underline{0.888} & \underline{28.06}/\underline{0.799} & \underline{36.37}/\underline{0.972} & \underline{30.58}/\underline{0.974} & \underline{32.06}/\underline{0.913} \\
\hline
  MTAIR(Ours) & \textbf{34.14}/\textbf{0.936} & \textbf{31.50}/\textbf{0.893} & \textbf{28.24}/\textbf{0.805} & \textbf{39.15}/\textbf{0.984} & \textbf{31.34}/\textbf{0.983} & \textbf{32.87}/\textbf{0.920} \\
\hline
\end{tabular}
\end{table*}

\begin{figure*}[t]
  \centering
  \includegraphics[width=\linewidth]{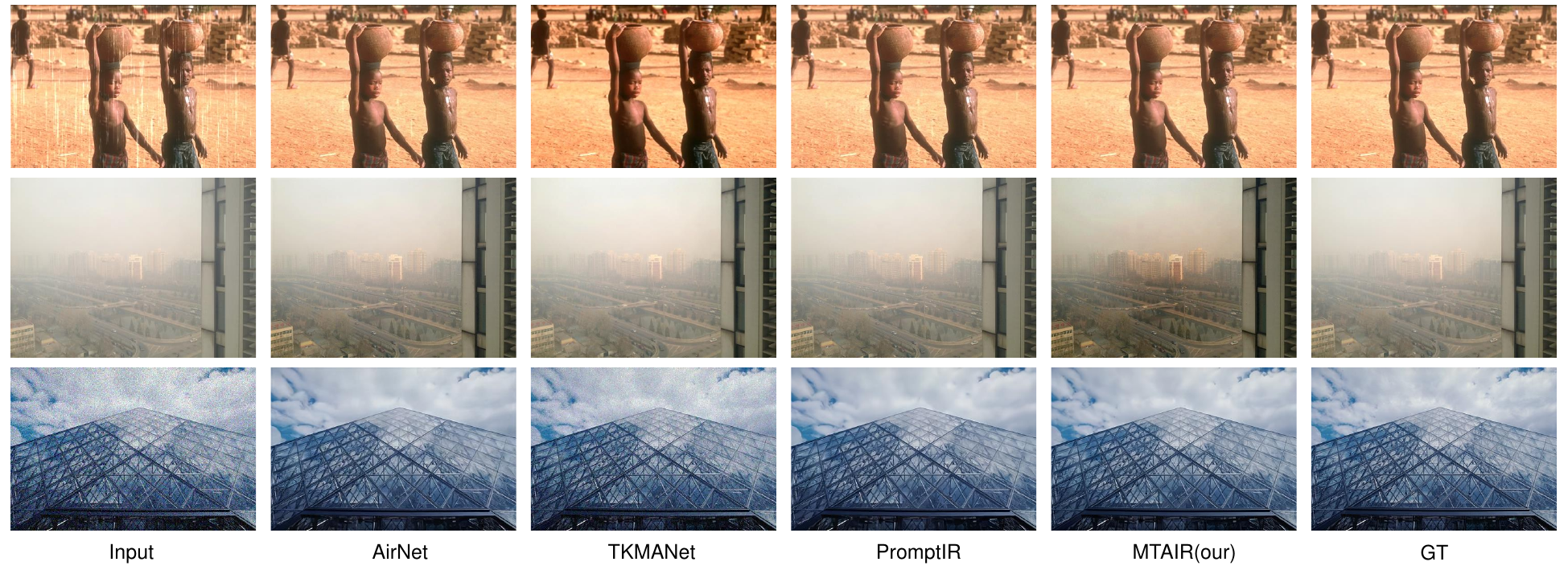}
  \caption{Visual comparisons with SOTA all-in-one models on Rain100L\cite{yang2017deep}, SOTS\cite{li2018benchmarking} and CBSD68\cite{martin2001database} sample images. The proposed model exhibits better degradation removal.}
  \label{fig:All-in-one result}
\end{figure*}

\begin{table*}[!ht]
\small 
\centering
\setlength{\tabcolsep}{9.0pt} 
\renewcommand\arraystretch{1.15} 
\caption{single task:denoise result. Comparisons with state-of-the-art image restoration methods under one-by-one restoration setting.}
\label{table:denoise result}
\begin{tabular}{c|ccc|ccc}
\hline
\multirow{2}{*}{Method} & \multicolumn{3}{|c}{ Denoising on CBSD68 dataset } & \multicolumn{3}{|c}{Denoising on Urban100 dataset} \\
  &  $\sigma=15$  &  $\sigma=25$  &  $\sigma=50$  &  $\sigma=15$  &  $\sigma=25$  &  $\sigma=50$ \\
\hline CBM3D\cite{dabov2007color} & 33.50/0.922 & 30.69/0.868 & 27.36/0.763 & 33.93/0.941 & 31.36/0.909 & 27.93/0.840 \\
  DnCNN\cite{zhang2017beyond} & 33.89/0.930 & 31.23/0.883 & 27.92/0.789 & 32.98/0.931 & 30.81/0.902 & 27.59/0.833 \\
  IRCNN\cite{zhang2017learning} & 33.87/0.929 & 31.18/0.882 & 27.88/0.790 & 27.59/0.833 & 31.20/0.909 & 27.70/0.840 \\
  FFDNet\cite{zhang2018ffdnet} & 33.87/0.929 & 31.21/0.882 & 27.96/0.789 & 33.83/0.942 & 31.40/0.912 & 28.05/0.848 \\
  BRDNet\cite{tian2020image} & 34.10/0.929 & 31.43/0.885 & 28.16/0.794 & 34.42/0.946 & 31.99/0.919 & 28.56/0.858 \\
  AirNet\cite{li2022all} & 34.14/0.936 & 31.48/0.893 & 28.23/0.806 & 34.40/0.949 & 32.10/0.924 & 28.88/0.871 \\
  Restormer\cite{zamir2022restormer} & 34.29/0.937 & 31.64/0.895 & 28.41/0.810 & 34.67/\underline{0.969} & 32.41/0.927 & 29.31/0.878 \\
  PromptIR\cite{potlapalli2024promptir} & \underline{34.34}/\underline{0.938} & \underline{31.71}/\underline{0.897} & \underline{28.49}/\underline{0.813} & \underline{34.77}/{0.952} & \underline{32.49}/\underline{0.929} & \underline{29.39}/\underline{0.881} \\
\hline
MTAIR(Ours) & \textbf{34.38}/\textbf{0.938} & \textbf{31.72}/\textbf{0.897} & \textbf{28.49}/\textbf{0.813} & \textbf{34.97}/\textbf{0.953} & \textbf{32.68}/\textbf{0.931} & \textbf{29.48}/\textbf{0.883} \\
\hline
\end{tabular}
\end{table*}

\begin{table*}[!ht]
\small 
\centering
\setlength{\tabcolsep}{3.0pt} 
\renewcommand\arraystretch{1.15} 
\caption{single task:derain result. Comparisons with state-of-the-art image restoration methods under one-by-one restoration setting.}
\label{tab:derain result}
\begin{tabular}{c|cccccccc}
\hline  \multirow{2}{*}{Method}& \multirow{2}{*}{MSPFN\cite{jiang2020multi}} & \multirow{2}{*}{LPNet\cite{gao2019dynamic}} & \multirow{2}{*}{MPRnet\cite{zamir2021multi}} & \multirow{2}{*}{AirNet\cite{li2022all}} & \multirow{2}{*}{Restormer\cite{zamir2022restormer}} & \multirow{2}{*}{TKMANet\cite{chen2022learning}} & \multirow{2}{*}{PromptIR\cite{potlapalli2024promptir}} & \multirow{2}{*}{MATIR(our)}\\
\\
\hline PSNR & 33.50 & 33.61 & \underline{38.26} & 34.90 & 36.74 & 35.60 & 37.04 & \textbf{39.27} \\
SSIM & 0.948 & 0.958 & \underline{0.982} & 0.977 & 0.978 & 0.974 & 0.979 & \textbf{0.985} \\
\hline
\end{tabular}
\end{table*}

\begin{table*}[!ht]
\small 
\centering
\setlength{\tabcolsep}{3.0pt} 
\renewcommand\arraystretch{1.15} 
\caption{single task:dehaze result. Comparisons with state-of-the-art image restoration methods under one-by-one restoration setting.}
\label{tab:dehaze result}
\begin{tabular}{c|cccccccc}
\hline  \multirow{2}{*}{Method}& \multirow{2}{*}{DehazeNet\cite{cai2016dehazenet}} & \multirow{2}{*}{EPDN\cite{qu2019enhanced}} & \multirow{2}{*}{MPRnet\cite{zamir2021multi}} & \multirow{2}{*}{AirNet\cite{li2022all}} & \multirow{2}{*}{Restormer\cite{zamir2022restormer}} & \multirow{2}{*}{TKMANet\cite{chen2022learning}} & \multirow{2}{*}{PromptIR\cite{potlapalli2024promptir}} & \multirow{2}{*}{MATIR(our)}\\
\\
\hline PSNR & 22.46 & 22.57 & 28.21 & 23.18 & 30.87 & \underline{31.37} & 31.31 & \textbf{31.70} \\
SSIM & 0.851 & 0.863 & 0.967 & 0.900 & 0.969 & \underline{0.974} & 0.973 & \textbf{0.983} \\
\hline
\end{tabular}
\end{table*}

\begin{figure*}[t]
  \centering
  \includegraphics[width=\linewidth]{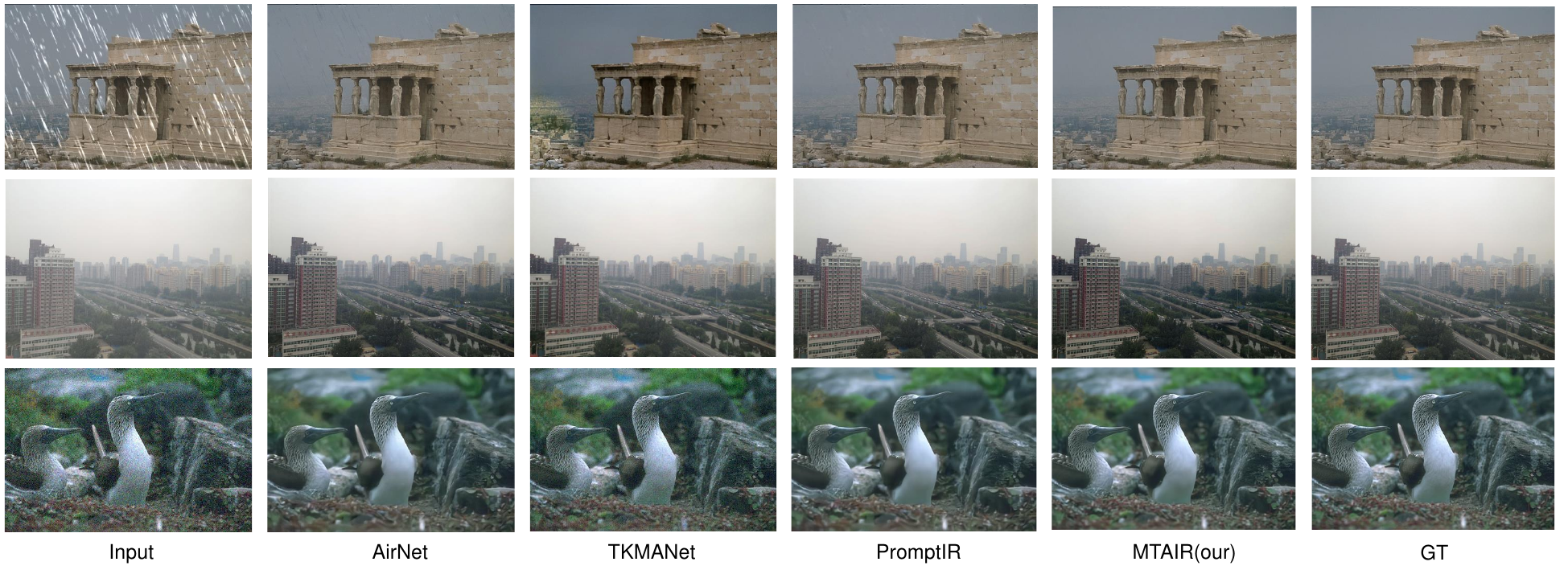}
  \caption{ Visual comparisons with SOTA models under single-task conditions on Rain100L\cite{yang2017deep}, SOTS\cite{li2018benchmarking} and CBSD68\cite{martin2001database} sample images. The proposed model exhibits better degradation removal.}
  \label{fig:single task result}
\end{figure*}

\subsection{Datasets}
For denoising task, datasets such as BSD400 \cite{martin2001database}, CBSD68 \cite{martin2001database}, WED \cite{ma2016waterloo}, and Urban100 \cite{huang2015single}, are utilized. The BSD400 dataset contains 400 clear images. CBSD68 contains 68 clear images, Urban100 contains 100 clear images, and WED contains 4744 clear images. Following general experiment settings in \cite{tian2020image,zhang2017beyond,zhang2017learning,zhang2018ffdnet}, we take images in both WED and BSD400 as training set, and images in Urban100 and CBSD68 as testing set. Three levels of Gaussian noise \(\sigma = \{15, 25, 50\}\) are added to these clear images to generate corresponding noisy images for training and evaluation.

For deraining task, dataset such as Rain100L \cite{yang2017deep} is utilized. It contains 200 pairs of rainy images for training, and 100 image pairs of testing.

For dehazing task, dataset such as the SOTS dataset in RESIDE \cite{li2018benchmarking} is utilized. It contains 72135 image pairs for training, and 500 image pairs for testing.

\subsection{Implementation details}
The numbers of M-T DHBlocks in the proposed MTAIR are set to be respective [4, 6, 6, 8] ranging from scale Level1 to Level4. Their respective channel numbers are set to be [48, 96, 192, 384]. The numbers of attention heads are accordingly set to be [1, 2, 4, 8].

All experiments are conducted using PyTorch on single NVIDIA RTX A5000 GPU. Adam optimizer \cite{kingma2014adam} with parameters (\(\beta_1 = 0.9\), \(\beta_2 = 0.999\) and weight decay \(1 \times 10^{-4}\)) is adopted. The initial learning rate is set to be \(2 \times 10^{-4}\). During training, batch size is set to be 8. Additionally, random horizontal and vertical flips are applied on training images for data augmentation. The images are cropped into patches of size 128 × 128 for training.

\subsection{Evaluation metrics}
Following previous works \cite{dong2020fd, fu2019lightweight, tian2020image}, we employ Peak Signal-to-Noise Ratio (PSNR) \cite{huynh2008scope} and Structural Similarity (SSIM) \cite{wang2004image} as our quantitative evaluation metrics.

In all performance tables, the best and the second-best performances are highlighted in bold and underlined, respectively. To further demonstrate the effectiveness of MTAIR, we have conducted experiments under both all-in-one and individual task settings.

\begin{table*}[!ht]
\small 
\centering
\setlength{\tabcolsep}{9.0pt} 
\renewcommand\arraystretch{1.15} 
\caption{Ablation experiment:M-T DHB. w/o  represents that the module is deleted. The best results are highlighted in bold.}
\label{Ablation experiment:M-T DHB}
\begin{tabular}{c|ccccc|c}
\hline
\multirow{2}{*}{model} & \multicolumn{3}{|c}{ Denoising on CBSD68 dataset } & {Deraining} & {Dehazing} & \multirow{2}{*}{ Average } \\
    &  $\sigma=15$  &  $\sigma=25$  &  $\sigma=50$  & on Rain100L & on SOTS  &     \\
\hline MTAIR w/o SSM & 34.10/0.934 & 31.45/0.891 & 28.17/0.801 & 38.82/0.983 & 31.03/0.980 & 32.71/0.917 \\
  MTAIR w/o CA & 34.12/0.935 & 31.46/0.891 & 28.21/0.803 & 38.91/0.983 & 31.12/0.980 & 32.76/0.918 \\
  MTAIR w/o M-T DIM & 34.11/0.934 & 31.48/0.891 & 28.21/0.801 & 39.05/0.983 & 31.15/0.981 & 32.80/0.918 \\
\hline MTAIR & \textbf{34.14}/\textbf{0.936} & \textbf{31.50}/\textbf{0.893} & \textbf{28.24}/\textbf{0.805} & \textbf{39.15}/\textbf{0.984} & \textbf{31.34}/\textbf{0.983} & \textbf{32.87}/\textbf{0.920} \\
\hline
\end{tabular}
\end{table*}

\begin{table*}[!ht]
\small 
\centering
\setlength{\tabcolsep}{9.0pt} 
\renewcommand\arraystretch{1.15} 
\caption{Ablation experiment:S-C PB. w/o  represents that the module is deleted. The best results are highlighted in bold.}
\label{tab:Ablation experiment:S-C PB}
\begin{tabular}{c|ccccc|c}
\hline
\multirow{2}{*}{model} & \multicolumn{3}{|c}{ Denoising on CBSD68 dataset } & {Deraining} & {Dehazing} & \multirow{2}{*}{ Average } \\
    &  $\sigma=15$  &  $\sigma=25 $ &  $\sigma=50$  & on Rain100L & on SOTS &         \\
\hline MTAIR w/o Spatial Prompt & 34.12/0.935 & 31.46/0.891 & 28.21/0.803 & 38.91/0.983 & 31.12/0.980 & 32.76/0.918 \\
  MTAIR w/o Channel Prompt & 34.13/0.935 & 31.48/0.892 & 28.23/0.805 & 39.07/0.984 & 30.93/0.981 & 28.77/0.719 \\
  MTAIR w/o S-C PIM & 34.12/0.935 & 31.47/0.892 & 28.20/0.805 & 39.03/0.984 & 31.11/0.981 & 32.78/0.919 \\
\hline MTAIR & \textbf{34.14}/\textbf{0.936} & \textbf{31.50}/\textbf{0.893} & \textbf{28.24}/\textbf{0.805} & \textbf{39.15}/\textbf{0.984} & \textbf{31.34}/\textbf{0.983} & \textbf{32.87}/\textbf{0.920} \\
\hline
\end{tabular}
\end{table*}

\begin{table*}[!ht]
\small 
\centering
\setlength{\tabcolsep}{9.0pt} 
\renewcommand\arraystretch{1.15} 
\caption{Ablation of degradation combinations.  ”\checkmark” represents MTAIR for corresponding degradation combination case, ”-” denotes unavailable results.}
\label{tab:Ablation of degradation combinations}
\begin{tabular}{ccc|ccc|c|c}
\hline
\multicolumn{3}{c|}{ Degration } & \multicolumn{3}{|c|}{ Denoising on CBSD68} &
\multirow{2}{*}{
\begin{tabular}{c}
Deraining \\
on Rain100L
\end{tabular}} & \multirow{2}{*}{\begin{tabular}{c}
Dehazing \\
on SOTS
\end{tabular}} \\
Noise & Rain & Haze &  \(\sigma=15\)  & \(\sigma=25\)  &  \(\sigma=50\)  & & \\
\hline \checkmark  & & & 34.38/0.938 & 31.72/0.897 & 28.49/0.813 & - & - \\
 &  \checkmark  & & - & - & - & 39.27/0.985 & - \\
 & &  \checkmark  & - & - & - & - & 31.70/0.983 \\
 \checkmark  &  \checkmark  & &  34.18 / 0.936  &  31.54 / 0.894  &  28.27 / 0.805  &  39.21 / 0.985  & - \\
 \checkmark  & &  \checkmark  &  34.12 / 0.934  &  31.47 / 0.891  &  28.20 / 0.803  & - &  31.02 / 0.980  \\
 &  \checkmark  &  \checkmark  & - & - & - &  38.60 / 0.982  &  31.18 / 0.980  \\
\hline \checkmark  &  \checkmark  &  \checkmark  &  34.14 / 0.936  &  31.50 / 0.893  &  28.24 / 0.805  &  39.15 / 0.984  &  31.34 / 0.983  \\
\hline
\end{tabular}
\end{table*}

\subsection{Comparison results on all-in-one task}
In this section, we primarily evaluate the performance of MTAIR on all-in-one task. To demonstrate its effectiveness, we compare the proposed method with several popular state-of-the-art methods. We selected four single-degradation image restoration methods (i.e., BRDNet \cite{tian2020image}, LPNet \cite{fu2019lightweight}, FDGAN \cite{dong2020fd}, and MPRNet \cite{zamir2021multi}) and six multi-degradation image restoration methods (i.e. DL \cite{fan2019general}, TKMANet \cite{chen2022learning}, AirNet \cite{li2022all}, DA-CLIP \cite{luo2023controlling}, LoRA-IR\cite{ai2024lora} and PromptIR \cite{potlapalli2024promptir}).

To ensure fair and accurate comparison, the training and testing settings are kept the same as those of the compared methods. From Table \ref{tab:all in one result}, we can observe that MTAIR outperforms almost all models. Some representative visual comparisons are illustrated in Figure~\ref{fig:All-in-one result}. From these visual results, we can observe that the proposed model can recover clear image with better visual qualities in color layering, detail richness, etc.

\subsection{Comparison results on individual tasks}
In this section, we primarily evaluate the performance of MTAIR on individual tasks. Individual models are trained for each corresponding restoration task. Specifically, from Table \ref{table:denoise result}, Table \ref{tab:derain result} and Table~\ref{tab:dehaze result}, we can observe that MTAIR outperforms almost all models on denoising, deraining, and dehazing task. Some representative visual comparisons are illustrated in Figure~\ref{fig:single task result}. From these visual results, we can as well observe that the proposed model can recover clear image with better visual qualities in color layering, detail richness, etc.

The experimental results convincingly show that MTAIR not only achieves significant improvements in all-in-one task, but also demonstrates competitive advantages in single-task mode compared to state-of-the-art methods.

\subsection{Ablation studies}
In this section, we have conducted several ablation experiments to analyze the impact of several key components on model's performance.

\subsubsection{Impact of M-T DHB}
We removed Vision State-Space Module (SSM), Channel Attention Module (CA), and M-T DIM in M-T DHB individually. As shown in Table \ref{Ablation experiment:M-T DHB}, it can be observed that ablation models with these modules removed do not achieve optimal results. It demonstrates that all these three integrated modules have positive effectiveness on model performance.

\subsubsection{Impact of S-C PB}
We removed Spatial Prompt, Channel Prompt, and S-C PIM in  S-C PB individually. As shown in Table \ref{tab:Ablation experiment:S-C PB}, it can be observed that ablation models with these modules removed do not achieve optimal results, which demonstrates the effectiveness of the Spatial Prompt, Channel Prompt, and S-C PIM.

\subsubsection{Impact of different degradation combinations on model performance}
We have evaluated the impact of different degradation type (task) combinations on the performance of MTAIR. The results on different combinations of all three restoration tasks are shown in Table \ref{tab:Ablation of degradation combinations}. From the results, we can observe that, as the number of degradation types increases, the network finds it increasingly difficult to restore clear images in all-in-one mode, leading to a little performance decline.

However, interestingly, we observe that the model trained on combination of rainy and noisy images achieved a little better performance than all-in-one three-tasks mode. Combining dehazing with deraining or denoising task resulted in a little worse performance than all-in-one three-tasks mode. It indicates positive correlation between deraining and denoising tasks, and negative influence on two other tasks. These observations raise an interesting question worth further exploration, which is beyond the scope of this paper.

\section{Conclusion}
In this paper, we have proposed an effective multi-dimensional visual prompt enhanced all-in-one image restoration model. By combining the modeling strengths of Mamba and Transformer, the proposed model can be implemented within restricted computation resource. Through introducing multi-prompt interaction modules in both spatial and channel directions, the proposed model have potentials to dynamically adjust feature distributions and mine correlated degradation properties through learnable prompts. Extensive experiments on public datasets have demonstrated that the proposed model achieves new state-of-the-art performance in typical image denoising, deraining, and dehazing tasks, when compared with many popular mainstream methods. Ablation studies have as well demonstrated the positive effectiveness of each key components.

\section*{Acknowledgment}
This work is supported by National Natural Science Foundation of China under Grand No. 62366021, and Jiangxi Provincial Graduate Innovation Funding Project under Grand YC2023-S298.

\ifCLASSOPTIONcaptionsoff
  \newpage
\fi

\bibliographystyle{IEEEtran}
\bibliography{myref}

\end{document}